\documentclass[]{interact}

\usepackage{epstopdf}
\usepackage[caption=false]{subfig}

\usepackage[longnamesfirst,sort]{natbib}
\bibpunct[, ]{(}{)}{;}{a}{,}{,}
\renewcommand\bibfont{\fontsize{10}{12}\selectfont}

\theoremstyle{plain}

\theoremstyle{definition}

\theoremstyle{remark}

\usepackage{xcolor}
\newcommand{\B}{\fontseries{b}\selectfont}

\begin{document}


\title{A hybrid optimization approach for employee rostering: Use cases at Swissgrid and lessons learned}

\author{
\name{Jangwon Park\textsuperscript{a}\thanks{Most of the work was carried out while Jangwon Park was with Swissgrid.} and Evangelos Vrettos\textsuperscript{b}\thanks{CONTACT Author: Evangelos Vrettos. Email: evangelos.vrettos@swissgrid.ch}}
\affil{\textsuperscript{a}University of Toronto, Canada; \textsuperscript{b}Research \& Digitalisation, Swissgrid Ltd., Aarau, Switzerland}
}

\maketitle

\begin{abstract} 
Employee rostering is a process of assigning available employees to open shifts. Automating it has ubiquitous practical benefits for nearly all industries, such as reducing manual workload and producing flexible, high-quality schedules. In this work, we develop a hybrid methodology which combines Mixed-Integer Linear Programming (MILP) with scatter search, an evolutionary algorithm, having as use case the optimization of employee rostering for Swissgrid, where it is currently a largely manual process. The hybrid methodology guarantees compliance with labor laws, maximizes employees' preference satisfaction, and distributes workload as uniformly as possible among them. Above all, it is shown to be a robust and efficient algorithm, consistently solving realistic problems of varying complexity to near-optimality an order of magnitude faster than an MILP-alone approach using a state-of-the-art commercial solver. Several practical extensions and use cases are presented, which are incorporated into a software tool currently being in pilot use at Swissgrid.
\end{abstract}


\section{Introduction}

Employee rostering aims to assign available employees to open shifts and is a common process in many industries, for example, healthcare, call and control centers, tourism, hospitality, and construction to name a few. The practical advantages of automated employee rostering compared with manual scheduling are widely accepted: easiness, efficiency, cost saving, compliance with labor laws and company directives, higher employee satisfaction, increased accountability and fairness, reduced under-staffing risk, and even improved health. 

Despite these benefits, only a limited number of commercially available software products perform automated employee rostering, as it remains a mathematically challenging problem. Indeed, most of the products aim at facilitating shift planning by means of an intuitive Graphical User Interface (GUI), but the shift assignment decisions are manually made by a human. To the authors' best knowledge, even the existing software products with true automation functionalities do not fully cover the business needs, use simplistic approaches or still require significant human involvement. As a result, many companies still perform employee rostering manually. 

To enable widespread use of automated employee rostering in the industry, software tools with two main properties are needed. First, a powerful and scalable solution engine capable of providing near-optimal shift schedules in due time. And second, a comprehensive and versatile application layer to support diverse company use cases.

In this direction, this paper presents an optimization-based hybrid methodology for employee rostering, which combines a Mixed-Integer Linear Programming (MILP) approach and a scatter search metaheuristic algorithm. The methodology is applied to the employee rostering problem at the control center of Swissgrid, the electricity transmission system operator of Switzerland. Swissgrid currently maintains a largely manual and time-consuming rostering process, which is rather inflexible in accommodating changes in employee work preferences or availability. Culminating in the implementation of a software tool with various supported applications, our work aims to bring the benefits of automated employee rostering at Swissgrid, but also provide useful insights for other industries.

Several approaches have been successful in solving real-life employee rostering problems, for example, nurse rostering. Table \ref{tab:Survey} summarizes a brief survey of studies categorized by algorithm, from which we can glean the following. 
\begin{itemize}
\item Pure, standard mathematical programming approaches (without decomposition) are scarce for modern problems due to their complexity. 
\item Single-solution-based metaheuristics are popular, but often supported by other problem-specific heuristics.
\item Evolutionary algorithms are generally successful without extensive use of problem-specific heuristics. 
\item The combination of mathematical programming and metaheuristics is generally less explored contrary to combinations of metaheuristics and problem-specific heuristics.
\end{itemize}

Therefore, in this paper, we propose a new approach to hybridize mathematical programming and an evolutionary algorithm called scatter search. The main innovation lies in using an MILP solver as a generator of feasible solutions that are used to initialize the scatter search population. Besides the substantial computational benefits, especially for large-scale problems, the hybrid methodology enables quantifying and monitoring the suboptimality of candidate solutions along the way. Further contributions include the integration of a heuristic relaxation step in mathematical programming and several enhancements to the standard scatter search algorithm.

In addition to the methodological innovation, our approach distinguishes itself from the existing scientific literature and commercial products by supporting a wide range of practical functionalities, including the following.  
\begin{itemize}
\item A mechanism to optimally adjust pre-computed rosters to unforeseen changes such as short-term unavailabilities.
\item Scaling up employee rostering to one year with an adaptive rolling horizon scheme that ensures uniform workload over time.
\item Incorporation of company-defined work patterns in employee rostering to reflect company
priorities in rostering.
\end{itemize}

\begin{table}[h]
\tbl{Brief survey of successful applications of optimization in employee rostering. References with an asterisk also use other problem-specific heuristics.}{
    \begin{tabular}{l l} \toprule
        Reference & Approach \\
    \midrule
    \multicolumn{2}{c}{Mathematical programming} \\
        Al-Yakoob and Sherali (2007) & Mixed-integer programming \\
        Kassa and Tizazu (2013) \\
        Agrali, Taskin, \& Unal (2017) \\
    \midrule
    \multicolumn{2}{c}{Single solution-based metaheuristics} \\
        Burke, De Causmaecker, \& Vanden Berghe (1999)$^*$ & Tabu search/local search\\
        Li, Lim, \& Rodrigues (2003) \\
        Meisels and Schaerf (2003)$^*$ & \\
        Burke, Kendall, \& Soubeiga (2003)$^*$ \\
        Xie, Potts, \& Bektas (2017) & Iterated local search \\
        Burke, Curtois, Post, Qu, \& Veltman (2008)$^*$ & Variable neighbourhood search \\
    \midrule
    \multicolumn{2}{c}{Population-based metaheuristics} \\
        Brezulianu, Fira, \& Fira (2009) & Genetic algorithm\\
        Zolfaghari, Quan, El-Bouri, \& Khashayardoust (2010) \\
        Tahanian and Khaleghi (2015) \\
        Maenhout and Vanhoucke (2010) & Scatter search \\
        Burke, Curtois, Qu, \& Vanden Berghe (2010) \\
    \midrule
    \multicolumn{2}{c}{Mathematical programming + metaheuristics} \\
        Li, Burke, Curtois, Petrovic, \& Qu (2012) & Falling tide algorithm \\
    \bottomrule
    \end{tabular}}
    \label{tab:Survey}
\end{table}

The paper is structured as follows. Section \ref{ProbDesc} describes the employee rostering problem at Swissgrid. In Section \ref{HybridMeth}, we present the details of the hybrid methodology along with our implementation of scatter search, followed by a comparison with an MILP-alone approach in Section \ref{CompareSolutions}. Section \ref{UseCases} discusses the practical extensions and use cases of our employee rostering model for Swissgrid, and Section \ref{Conclusion} concludes our work.

\section{Problem description}\label{ProbDesc}
In employee rostering, we seek to assign available employees to open shifts while complying with labor regulations, company directives, and employees' personal requirements. In this section, we describe the general characteristics of the problem as well as aspects that may be specific to Swissgrid.

\subsection{General characteristics of employee rostering}
\subsubsection{Multi-shift environment and multi-skilled workforce}
Typically, there are multiple shift types and employees may have licenses to perform one or more of them. The \textit{license mix} within the workforce dictates who is legally entitled to perform which shift types, which is determined by appropriate compulsory training.

\subsubsection{Employee unavailabilities and preferences}
We consider employees' unavailabilities and preferences in optimizing their schedules. Unavailabilities can be short- or long-term: short-term unavailabilities typically range from a few hours to a few days, whereas long-term unavailabilities refer to vacation. While the number of vacation days is constant for everyone, the number of short-term unavailabilities may vary considerably. Preferences indicate if employees wish, or not wish, to work during certain time slots on a given day. Therefore, the problem formulation for employee rostering should forbid assigning employees to shifts during their unavailable times and violate employees' preferences as little as possible.

\subsubsection{Labor regulations}
Some common labor regulations, which also apply to employee rostering at Swissgrid, are outlined in Table \ref{tab:CommonLaborRegs}.

\begin{table}
\tbl{Commonly considered labor regulations in employee rostering}{
    \begin{tabular}{ll} \toprule
        \B{Regulation} & \B{Description} \\
        \midrule
        Minimum shift & Lowest allowable number of shifts worked per week. \\
        Maximum shift & Highest allowable number of shifts worked per week. \\
        Minimum rest days & Lowest allowable number of rest days in the given planning horizon. \\
        Maximum rest days & Highest allowable number of rest days in the given planning horizon. \\
        Minimum rest hours & Guaranteed rest hours between any two shifts. \\
        Minimum Sunday rest & Lowest guaranteed number of rest Sundays in the given planning horizon. \\
        License compliance & Each employee must have the appropriate license for the shift that they perform. \\
        \bottomrule
    \end{tabular}}
    \label{tab:CommonLaborRegs}
\end{table}

\subsection{Specific characteristics of Swissgrid}
\subsubsection{Cost of employee assignment}
In general, the costs of assigning employees typically vary due to contractual differences. At Swissgrid, however, we assume constant assignment costs in the problem formulation. Consequently, uniformizing workload distribution, and thus ensuring fairness across employees, is a critical component of the objective function.

\subsubsection{Shift types}\label{shifttypes}
The System Operations department at Swissgrid is responsible for operational planning and real-time control of the Swiss electricity transmission grid. This involves planning outages of grid assets for maintenance purposes, activation of power reserves to maintain grid stability, and switching operations for controllable grid components to optimize the flow of electricity. System operation actions are taken and coordinated by five teams at the two control centers of Swissgrid located at Aarau (German-speaking part of Switzerland) and Prilly (French-speaking part). In general, each team has different types and numbers of shifts, which are well-defined and fixed for each yearly operation, but might change once every few years according to the needs.


In this work, we focus on a team in Prilly which, at the time of writing, had three shift types: regional switching operations, on-call duty (P), and outage planning (OM). The duration of a shift can vary from type to type. For example, a switching operations shift lasts 8 hours and is thus assigned to employees as a morning (M), afternoon (A), or night (N) shift. An on-call duty and outage planning shift, however, takes a whole day.

\subsubsection{Forbidden shift type sequences}
For the considered team at Prilly, employees are forbidden to work a morning shift if an on-call duty was assigned to them on the previous day to ensure sufficient rest under all circumstances. In general, a multi-shift environment may come with a set of special relationships among different shift types that prohibit certain sequences.

\subsubsection{Cover requirements}\label{coverreq}
Different shift types have different \textit{cover requirements}, i.e., the number of employees required at any time. The cover requirements at Swissgrid are dictated by grid security considerations; therefore they are deterministic and need exact satisfaction. Key aspects of the shift types considered in this work are summarized in Table \ref{tab:ShiftTypesPrilly}.


\begin{table}
\tbl{Key aspects of the shift types considered in the problem formulation. There must be one employee every morning, afternoon, and night all year long including public holidays for switching operations and on-call duties. In contrast, outage planning is only conducted on business days. Note that weekends are treated specially at Swissgrid because morning and night shifts are extended to 12 hours, thus eliminating afternoon shifts entirely.}{
    \begin{tabular}{ccccc} \toprule
        \B{Shift type} & \B{Duration} & \B{Cover} & \B{Weekend cover} & \B{Holiday cover} \\
        \midrule
        Switching operations & 8 hours & 1 & Yes & Yes \\
        On-call duty (P) & All day & 1 & Yes & Yes \\
        Outage planning (OM) & All day & 1 & No & No \\
        \bottomrule
    \end{tabular}}
    \label{tab:ShiftTypesPrilly}
\end{table}

\subsubsection{Rest day}\label{restday}

\begin{figure}
    \centering
    \includegraphics[width=0.75\textwidth]{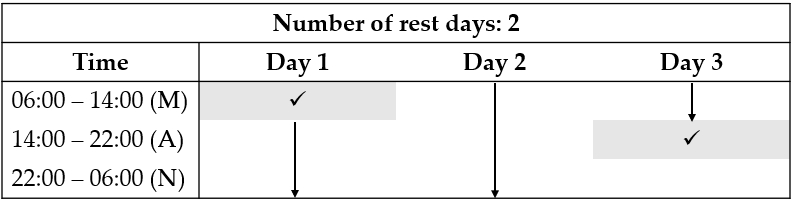}
    \includegraphics[width=0.75\textwidth]{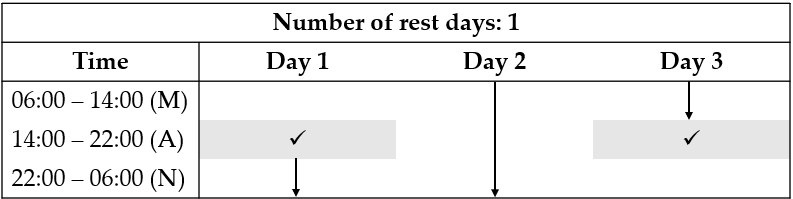}
    \caption{Examples of three-day schedules. In the top schedule, an employee has a morning shift on day 1 and an afternoon shift on day 3, which suggests two rest days, as there are 48 consecutive hours of non-shift time. However, changing the morning shift on day 1 to afternoon, as shown by the bottom schedule, results in only one rest day despite the same number of shifts. Note that if a rest day was defined in a simpler way as a full calendar day without a shift, which is nevertheless not Swissgrid's approach, then both the top and bottom schedule would correspond to one rest day.}
    \label{fig:RestDayEx}
\end{figure}

The number of rest days is calculated by counting the total instances of consecutive 24 hours of non-shift time throughout the planning horizon. Based on this definition, the number of rest days of two employees can be different despite having the same number of shifts. Figure \ref{fig:RestDayEx} provides a visual explanation of the sensitivity of rest days to shift arrangements.

\subsubsection{Planning horizon and short-term adjustments to pre-computed shift plans}
Providing an annual schedule in advance is highly desired at Swissgrid, as it allows time for long-term planning. Later in the year, however, employees may change their unavailabilities and preferences. For example, employees coordinate their vacation time with their families and may update their unavailabilities in the future, which may render the currently optimal annual plan suboptimal or even infeasible. Optimizing short-term adjustments to the pre-computed annual plan is therefore necessary.

\section{Hybrid solution methodology: MILP and scatter search}\label{HybridMeth}
\subsection{Overall methodology}
Our main decision variables are binary, and they represent whether or not an employee should work a given shift type at a given time slot. These decisions should minimize employees' preference violation and uniformize workload among them as best as possible. Our first solution approach is a
MILP, whose mathematical formulation is provided in Appendix A. 

The computational cost of solving the MILP increases dramatically with the problem size, e.g., number of employees. In fact, mathematical programming approaches may be inappropriate for the large, complex search spaces typically posed by modern rostering problems. Metaheuristics on the other hand are often preferred because they produce reasonably good solutions within limited computational time (Ernst, Jiang, Krishnamoorthy \& Sier, 2004). However, many metaheuristics do not explicitly measure the suboptimality of a feasible solution. Therefore, we develop a hybrid methodology that combines the MILP sequentially with scatter search -- a population-based evolutionary algorithm -- to overcome the weaknesses of either approach and solve larger problems more tractably. 

\begin{figure}
    \centering
    \includegraphics[width=1\textwidth]{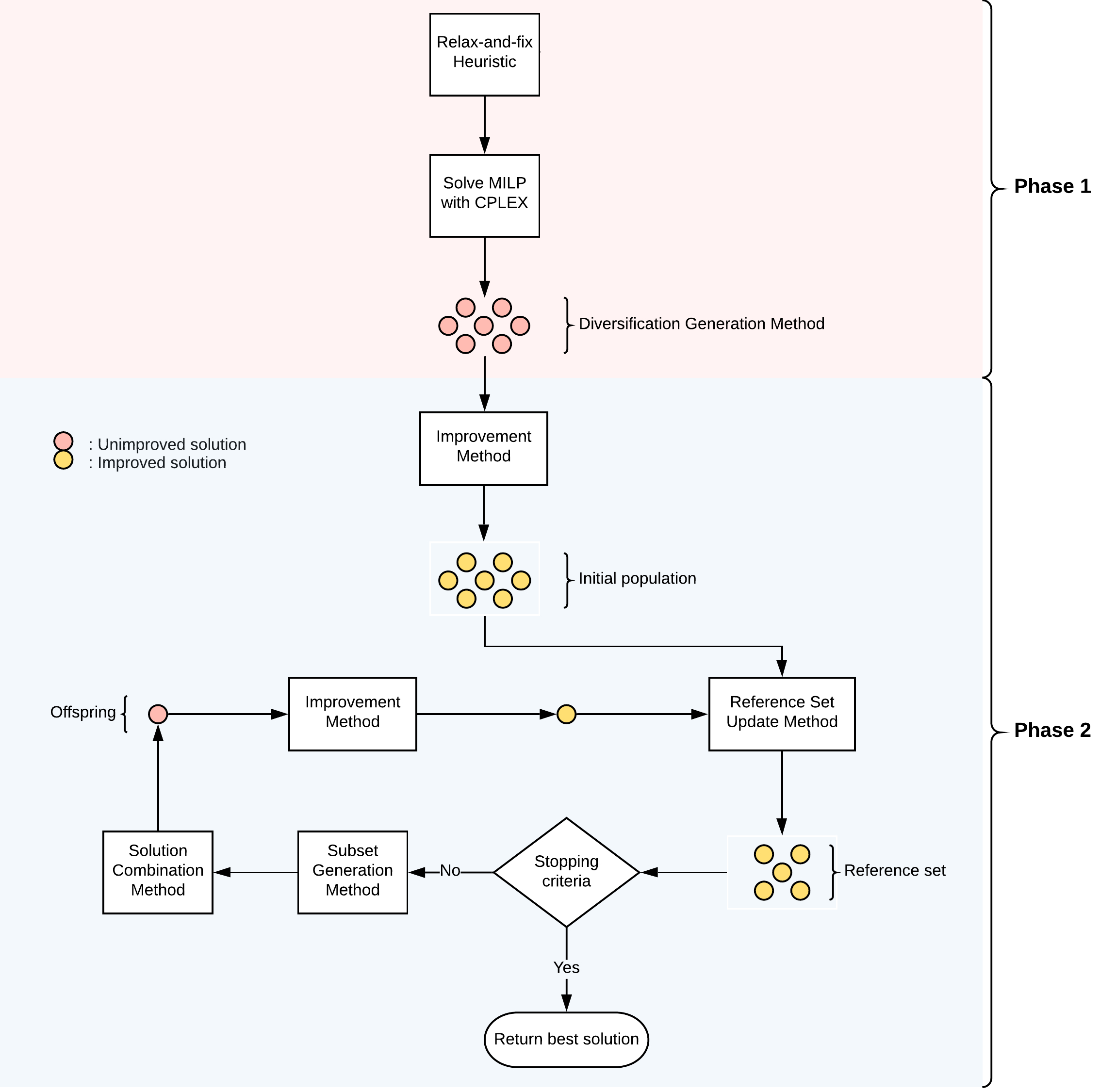}
    \caption{Flowchart of the hybrid methodology combining MILP and scatter search.}
    \label{fig:Flowchart}
\end{figure}

The proposed hybrid methodology is a two-phase algorithm, as illustrated in Figure \ref{fig:Flowchart}. In the first phase, we employ CPLEX to (1) solve the MILP for a set of feasible solutions and (2) extract a lower bound of the problem from its branch-and-bound tree. In the second phase, we employ scatter search to improve the feasible solutions until the \textit{optimality gap} of the best solution, i.e., the degree to which a solution is suboptimal, is sufficiently close to 0 (optimality). The desired optimality gap is determined by the user.

The first phase of the hybrid methodology is subdivided into two further steps. In the first step, all integrality constraints are relaxed, and we solve the resulting linear program to optimality. The values of the decision variables which turn out to be integral are fixed in the second step and we solve the resulting MILP. We call this pre-processing scheme the \textit{relax-and-fix heuristic}, and it offers two empirical advantages: (1) reducing the time spent to search for a set of feasible solutions, and (2) improving the quality of each solution in the set.

In the following sections, we summarize the implementation of the five subroutines of scatter search for the employee rostering problem at Swissgrid.

\subsection{Diversification generation method}
The diversification generation method creates a diverse set of feasible solutions. Unlike randomly generating an initial population of feasible solutions as in Burke, Curtois, Qu et al. (2010), we use CPLEX with user-defined parameters to search for a population of solutions. Additionally, we can control: (1) how much time is spent in this search, (2) how solutions already in the population are replaced, and (3) how much diversity the replacement strategy should consider. Therefore, CPLEX is a flexible alternative to the conventional diversification generation method. Note that similar possibilities exist with other commercial solvers (e.g., Gurobi), which would be equally suitable as the diversification generation method.

\subsection{Improvement method}
\begin{figure}
    \centering
    \includegraphics[width=1\textwidth]{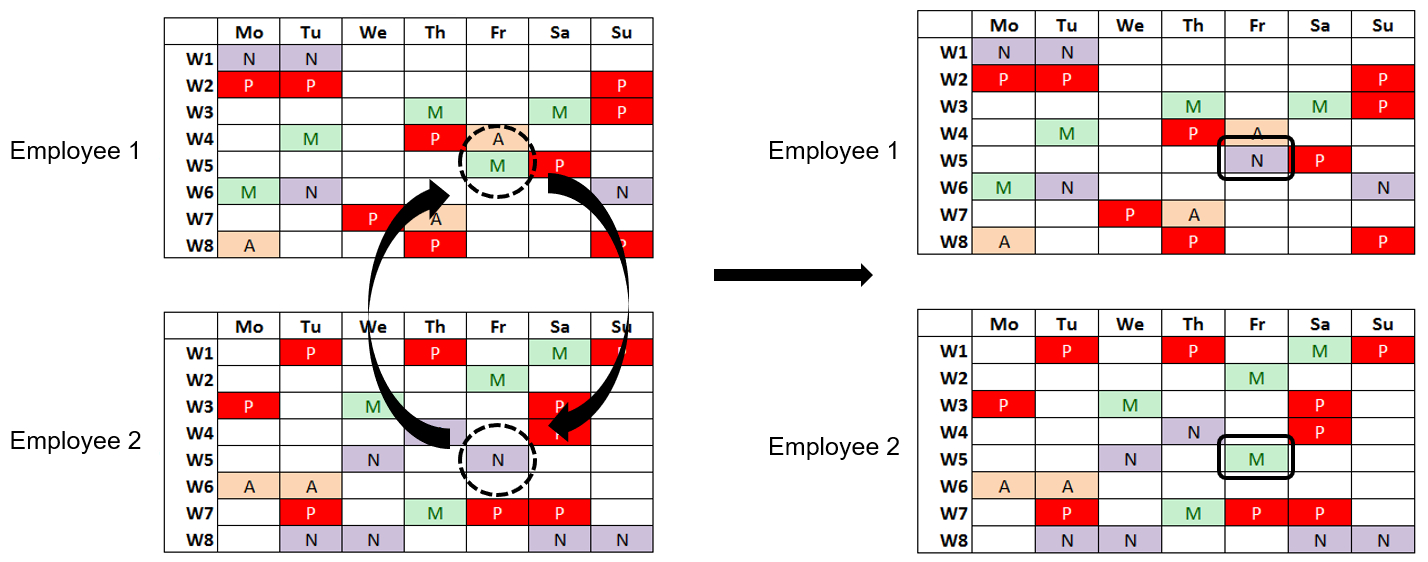}
    \caption{Swap of shifts between employees 1 and 2 on Friday of week 5 of the year, all of which are selected at random. Recall that ``M," ``A," and ``N" indicate morning, afternoon, and night shifts, respectively, whereas ``P" refers to an on-call duty.}
    \label{fig:LocalSearch}
\end{figure}

The improvement method attempts to improve the quality of a given solution. For this purpose, we use local search. Within local search, we consider two randomly selected employees for a \textit{swap} of shifts on a random date, as illustrated in Figure \ref{fig:LocalSearch}. If the swap improves the solution's objective value, it is accepted; otherwise, it is reversed. Afterwards, we proceed with another selection of employees and date. Local search terminates when the number of consecutive swaps that did not lead to an improvement reaches a threshold. 

In general, a higher threshold may achieve more improvements, but may also be more time-consuming. One strategy is to start with a relatively small threshold in the beginning, but increase it if the best solution of the current generation of scatter search is not better than that of the last generation by more than 10\%, for example. To make local search more effective, we select employees according to probabilities which are equal to the quality of each employee's schedule divided by the overall objective value. In other words, the employee with the worst schedule is the most probable one to be selected for a swap.

\subsection{Reference set update method}
The update method for the \textit{reference set}, i.e., the fixed-size population of solutions, used in this work follows the strategy by Burke, Curtois, Qu et al. (2010). A solution improved by local search replaces the worst solution in the reference set if it is strictly better and not already in the set. When the reference set has yet to reach its fixed size, any new solution will automatically enter the set.

\subsection{Subset generation method}
The subset generation method identifies which solutions from the reference set (parents) are used to create a new solution (offspring). Scatter search allows more than two parents to create an offspring.

The subset generation method in this work follows Burke, Curtois, Qu et al. (2010) closely:
\begin{enumerate}
    \item \textit{2-Subset}: Construct all unique subsets of the reference set (\textit{RefSet}) containing two solutions.
    \item \textit{n-Subset}: For every $n = 3, ..., |\textit{RefSet}|$, construct all unique subsets of size $n$ created by adding to each element in \textit{(n$-$1)-Subset} the best solution not already in the element.
\end{enumerate}

To avoid constructing a subset with no new solutions from the last generation of scatter search, each offspring is initially marked as ``new" when entering the next generation, but switched to ``old" in subsequent generations. Scatter search terminates when no new solutions are observed.

\subsection{Solution combination method}
The solution combination method creates an offspring from the parent solutions. Our approach is similar to that by Burke, Curtois, Qu et al. (2010), but has more randomized components. 


A solution is simply a collection of employee-shift-day triplets where ``no shift," or rest, is also considered a shift type in this context. A \textit{candidate} is defined to be a particular triplet, and the number of votes for the candidate is the number of parents that contain the candidate. In general, candidates with the most number of votes will populate the empty offspring first. \textit{Unanimous attributes}, i.e., candidates that all parent have, are inherited directly. Visually, this means that the part of the employee roster matrix (see Figure \ref{fig:LocalSearch}) that is identical in all combined parents will carry over to the offspring.

Ties occur between candidates with equal votes if they have the same employee and day, but not the shift. Burke, Curtois, Qu et al. (2010) uses a rule-based deterministic approach to break ties, which is reasonable because scatter search is their main methodology and they use a more complex improvement method called variable depth search. In contrast, we break ties by randomly selecting one of the candidates. This simpler stochastic approach is better suited to escape local optima in our case because scatter search is employed in conjunction with a deterministic MILP.

The offspring may still under-satisfy cover requirements after processing all votes and candidates. This is corrected by a simple heuristic. Specifically, we sort employees in increasing order of workload according to their current schedules and attempt to assign shifts to employees with the least number of shifts first.
This heuristic accelerates uniformizing the workload among employees, which is one of our objectives.

\section{Comparison of solution approaches}\label{CompareSolutions}
\subsection{Assumptions and parameters}
In this section, we compare the two solution approaches, namely, the MILP-alone approach and the hybrid methodology detailed throughout Section \ref{HybridMeth}. The list below describes the key assumptions and parameter settings used in this analysis. Although the assumptions are largely based on the situation of a Swissgrid team at the time of writing, they do not necessarily or thoroughly reflect the actual parameters, regulations and principles of employee rostering applied by Swissgrid.

\begin{itemize}
    \item The \textit{base problem instance} has 12 employees and three shift types, as described in Section \ref{shifttypes}. All employees perform switching operations, but half are entitled to perform on-call duties (P) and the other half outage planning (OM).
    \item Cover requirements are constant week over week and as described in Section \ref{coverreq} and Table \ref{tab:ShiftTypesPrilly}. Holidays are based on the year 2020. 
    \item If simulating with a number of employees other than 12, the cover requirements are scaled proportionally to the number of employees.
    \item All employees are available 95\% of the time; the exact days are randomized.
    \item All employees receive 25 days of vacation per year (separate from availability above).
    \item Employees express preferences up to 20\% of the slots in the planning horizon. This is roughly equal to 1 - 2 days per week. The exact days with preferences are randomized. Once sampled, it is a preference ``for" or ``against" working with a 50-50 chance. 
    \item All problem instances have an 8-week planning horizon, inspired by the considered Swissgrid team whose approach has been to plan 8 weeks first and repeat it for the rest of the year.
    \item Based on similar nurse rostering problems in Burke, Curtois, Qu et al. (2010), the size of the reference set within scatter search is five. \item The size of the initial population is set to six, which effectively allows us to eliminate the worst solution that our diversification method (MILP with CPLEX) finds by later pruning it.
\end{itemize}

Under these assumptions, the base problem instance has approximately 6,500 continuous and 8,000 integer variables, and nearly 40,000 constraints, excluding non-negativity and integrality restrictions. Besides the base problem instance, we also solve scaled-up versions of the problem with 24, 36, 48, 60, and 72 employees (with randomized employee availability and preference parameters). 

\subsection{Computational time comparison}\label{CompTimeCompare}
For each workforce size from 12 to 72 employees, we record the mean computational time for each approach to reach various optimality gaps across 5 randomized trials. The lower the mean computational time at a given optimality gap, the more efficient the solution approach.


Table \ref{TimeCompare} demonstrates that for problems with 24 employees or more, our hybrid methodology is more efficient than the MILP-alone approach by at least an order of magnitude in converging to optimality gaps 5\% or lower, which would be an acceptable solution quality in practice. While the MILP-alone approach is already impractical for problems with 48 employees (e.g., longer than 1 hour per 8-week planning horizon to reach 3\% optimality gap), the hybrid methodology remains practical up to and including the largest problem instance. With 72 employees, it allows us to solve to near-optimality in roughly 9 minutes on average, whereas the MILP-alone approach takes nearly 2.5 hours. This can be partly explained by the fact that while CPLEX tends to become ``stuck" at certain plateaus, scatter search continuously and rapidly finds better solutions throughout the algorithm. However, for small problems (12 employees), the solution time of the MILP-alone approach is smaller than the time required to execute the relax-and-fix heuristic of the hybrid methodology.

The results are similar even for different parametric assumptions. For example, assuming that employees provide preferences for 50\% of the planning horizon (roughly 3 - 4 days per week) does not appreciably change the observations. The hybrid methodology continues to outperform the MILP-alone approach by at least an order of magnitude in all problems with 24 employees or more and reaches the 1\% optimality gap. In fact, with sparser preference densities, the spread tends to become larger. To confirm this, we experimented with 0\% preferences (no preferences). We consequently found that the hybrid methodology is more efficient even in the smallest problems, as shown in Table \ref{ZeroPreferencesTime}. Moreover, the performance difference between the two approaches is much wider compared to the case where employees provide some preferences (Table \ref{TimeCompare}).
\begin{table}
\tbl{Comparison of mean computational time in seconds across 5 randomized trials, assuming up to 20\% preferences. The columns (12 to 72) represent the number of employees in the problem instances. Bold fonts indicate that the corresponding approach achieved a lower time than the other in solving the same problems. In the hybrid methodology, the optimality gaps can be computed using the best bound from CPLEX, obtained in Phase 1 of Figure \ref{fig:Flowchart}.}
{\begin{tabular}{ccccccc|cccccc} \toprule
  & \multicolumn{6}{c}{Hybrid methodology} & \multicolumn{6}{c}{MILP-alone approach} \\ \cmidrule{2-13}
  Optimality gap & 12 & 24 & 36 & 48 & 60 & 72 & 12 & 24 & 36 & 48 & 60 & 72 \\ \midrule
 50\%  & 165 & \B176 & \B223 & \B219 & \B256 & \B317 & \B83 & 403 & 2206 & 2785 & 5242 & 6540 \\
 20\% & 165 & \B176 & \B223 & \B219 & \B256 & \B317 & \B83 & 403 & 2206 & 2785 & 5242 & 6540 \\
 10\% & 165 & \B176 & \B223 & \B219 & \B257 & \B318 & \B83 & 403 & 2206 & 2785 & 5242 & 8097 \\
 5\% & 165 & \B176 & \B223 & \B222 & \B257 & \B331 & \B83 & 810 & 2310 & 3368 & 7950 & 8097 \\
 3\% & 165 & \B176 & \B223 & \B236 & \B318 & \B346 & \B94 & 995 & 3052 & 4395 & 9239 & 8791 \\
 1\% & 167 & \B183 & \B230 & \B382 & \B526 & \B517 & \B94 & 995 & 3052 & 4395 & 9239 & 9007 \\ \bottomrule
\end{tabular}}
\label{TimeCompare}
\end{table}

\begin{table}
\tbl{Comparison of mean computational time in seconds across 5 randomized trials, assuming 0\% preferences. The columns (12 to 72) represent the number of employees in the problem instances. Bold fonts indicate that the corresponding approach achieved a lower time than the other in solving the same problems.}
{\begin{tabular}{ccccccc|cccccc} \toprule
  & \multicolumn{6}{c}{Hybrid methodology} & \multicolumn{6}{c}{MILP-alone approach} \\ \cmidrule{2-13}
  Optimality gap & 12 & 24 & 36 & 48 & 60 & 72 & 12 & 24 & 36 & 48 & 60 & 72 \\ \midrule
 50\%  & 171 & 256 & \B567 & \B1160 & \B1424 & \B2074 & \B111 & \B146 & 920 & 1884 & 5432 & 6349 \\
 20\% & 171 & \B256 & \B567 & \B1160 & \B1424 & \B2074 & \B152 & 2315 & 3753 & 17136 & 43336 & 66539 \\
 10\% & 171 & \B256 & \B567 & \B1160 & \B1424 & \B2074 & \B156 & 2315 & 3753 & 17136 & 43336 & 66539 \\
 5\% & \B171 & \B256 & \B567 & \B1160 & \B1424 & \B2076 & 251 & 2315 & 3753 & 20316 & 79728 & 66539 \\
 3\% & \B171 & \B256 & \B567 & \B1160 & \B1424 & \B2076 & 360 & 2691 & 7910 & 20316 & 81243 & $>$ 24 h* \\
 1\% & \B171 & \B260 & \B572 & \B1167 & \B1496 & \B2145 & 360 & 2691 & 7910 & 20316 & 81243 & $>$ 24 h* \\ \bottomrule
\end{tabular}}
\tabnote{\textsuperscript{*}Computational times were measured only up to 24 hours.}
\label{ZeroPreferencesTime}
\end{table}

\begin{figure}
    \centering
    \includegraphics[width=1\textwidth]{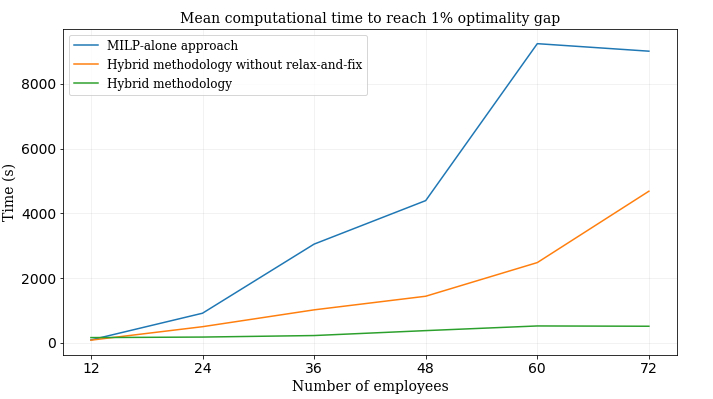}
    \caption{Comparison of mean computational time to reach  1\% optimality gap for various problem instances.}
    \label{fig:ScaleCompare}
\end{figure}

The final row of Table \ref{TimeCompare} offers additional insights into the scalability of the two solution approaches with respect to the number of employees. Figure \ref{fig:ScaleCompare} exhibits these values and highlights the increasing gap between the mean computational time of the MILP-alone approach and the hybrid methodology, which clearly indicates that the latter is more scalable with respect to workforce size.

Furthermore, Figure \ref{fig:ScaleCompare} shows that the computation time grows faster without the relax-and-fix heuristic (without the first step in Figure \ref{fig:Flowchart}). The relax-and-fix heuristic is useful for reducing the overall computational time because our decision variables and parameters are mostly binary-coded. Indeed, we observed that most decision variables turned out to be integers in the optimal solution to the linear-relaxed problem. Fixing these variables reduces many degrees of freedom, and consequently time, in searching for an initial population of solutions.

In Table \ref{HybridsCompare}, we further observe that the relax-and-fix heuristic provides scatter search with an advanced starting point, i.e. a much better initial population. For instance, using the relax-and-fix heuristic, we hardly observe any difference in the time to reach 50\% optimality gap relative to the time to reach 10\%; the same, however, cannot be said without the heuristic.
\begin{table}
\tbl{Comparison of mean computational time in seconds across 5 randomized trials between two variants (with and without the relax-and-fix heuristic) of the hybrid methodology.}
{\begin{tabular}{ccccccc|cccccc} \toprule
  & \multicolumn{6}{c}{With relax-and-fix heuristic} & \multicolumn{6}{c}{Without relax-and-fix heuristic} \\ \cmidrule{2-13}
  Optimality gap & 12 & 24 & 36 & 48 & 60 & 72 & 12 & 24 & 36 & 48 & 60 & 72 \\ \midrule
 50\%  & 165 & 176 & 223 & \B219 & \B256 & \B317 & \B87 & \B117 & \B158 & 249 & 323 & 597 \\
 20\% & 165 & 176 & 223 & \B219 & \B256 & \B317 & \B87 & \B118 & \B159 & 253 & 331 & 611 \\
 10\% & 165 & 176 & \B223 & \B219 & \B257 & \B318 & \B87 & \B161 & 260 & 449 & 562 & 840 \\
 5\% & 165 & \B176 & \B223 & \B222 & \B257 & \B331 & \B87 & 234 & 399 & 770 & 954 & 1376 \\
 3\% & 165 & \B176 & \B223 & \B236 & \B318 & \B346 & \B87 & 266 & 494 & 911 & 1246 & 1903 \\
 1\% & 167 & \B183 & \B230 & \B382 & \B526 & \B517 & \B87 & 503 & 1023 & 1443 & 2482 & 4685 \\ \bottomrule
\end{tabular}}
\label{HybridsCompare}
\end{table}


\subsection{Robustness comparison}
The standard deviation of the optimality gaps from randomized trials reveals whether a solution approach is robust. Specifically, a low standard deviation suggests that the approach can find roughly equally qualified solutions irrespective of the randomized availability and preference parameters, whereas a high standard deviation implies that the effectiveness of the approach depends heavily on the parameters. A practical employee rostering tool should consistently produce solutions of similar qualities by a specified time limit, for example, 5, 10, 15, or 30 minutes.

Results in Table \ref{RobustCompare} are reported for 50\% preferences, i.e. employees can specify up to 50\% of the slots in the planning horizon for preferences. If less than 50\%, the MILP-alone approach cannot produce any feasible solution within 30 minutes for problems with more than 24 employees. Even for 50\% preferences, the MILP-alone approach again cannot generate any feasible solution within 30 minutes beyond 48 employees, which explains why we report numbers only up to 48 employees.

Table \ref{RobustCompare} shows that the standard deviation of the optimality gaps at each time limit can be very large and inconsistent for the MILP-alone approach. On the contrary, the standard deviations from the hybrid methodology are relatively low for all problems at each time limit. Furthermore, the standard deviation tends to decline with time; the more time we allow, the more similar solution qualities become. The same, however, cannot be observed with the MILP-alone approach which produces standard deviations that can either increase or decrease with time.
\begin{table}
\tbl{Comparison of standard deviation of optimality gaps across 5 randomized trials. The columns (24, 36 and 48) represent the number of employees in the problem instances. The problem instance with 12 employees is omitted because both the MILP-alone approach and the hybrid methodology reach $<$ 1\% optimality gap before 5 minutes.}
{\begin{tabular}{cccc|ccc} \toprule
  & \multicolumn{3}{c}{Hybrid methodology} & \multicolumn{3}{c}{MILP-alone approach} \\ \cmidrule{2-7}
  Time elapsed (min) & 24 & 36 & 48 & 24 & 36 & 48 \\ \midrule
 5  & 0.5\% & 1.0\% & 2.9\% & 18.9\% & 21.2\% & 1.6\% \\
 10 & 0.2\% & 0.8\% & 1.5\% & 1.6\% & 20.9\% & 2.6\% \\
 15 & * & 0.5\% & 0.9\% & * & 20.9\% & 2.6\% \\
 30 & * & * & 0.2\% & * & 1.6\% & 18.2\% \\ \bottomrule
\end{tabular}}
\tabnote{\textsuperscript{*}By this point, 1\% optimality gap has been reached in all 5 randomized trials, and the standard deviation does not matter as much.}
\label{RobustCompare}
\end{table}

\begin{figure}
    \centering
    \includegraphics[width=0.7\textwidth]{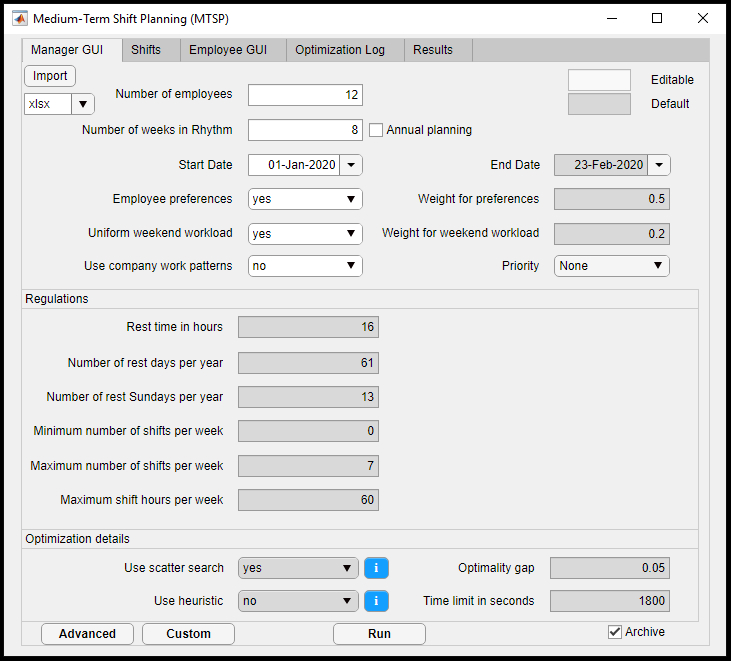}
    \caption{Among other things, the manager GUI is used to specify the number of employees, the planning horizon, and how to weigh employee preferences and uniform workload in the objective. Various labor regulations and optimization details (e.g., solution methodology and optimality gap) are also specified.}
    \label{fig:ManagerGUI}
\end{figure}




\begin{figure}
    \centering
    \includegraphics[width=0.7\textwidth]{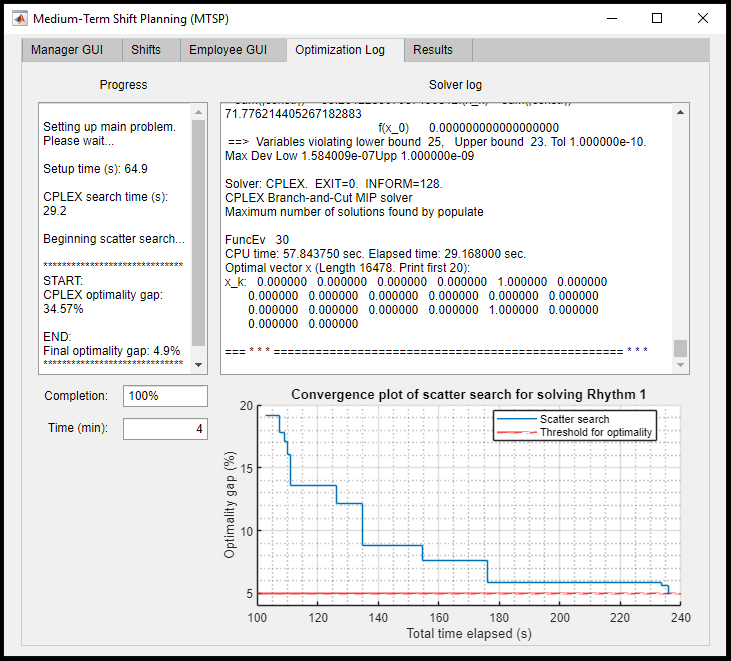}
    \caption{The solution progress is updated in real time including a convergence plot to monitor algorithm performance as it is approaching the desired solution quality. In this example, CPLEX terminated with an optimality gap of 34.57\%, whereas scatter search improved the solution to an optimality gap of 4.9\% (i.e., below the acceptance threshold of 5\%).}
    \label{fig:OptimizationLog}
\end{figure}

\begin{figure}
    \centering
    \includegraphics[width=1\textwidth]{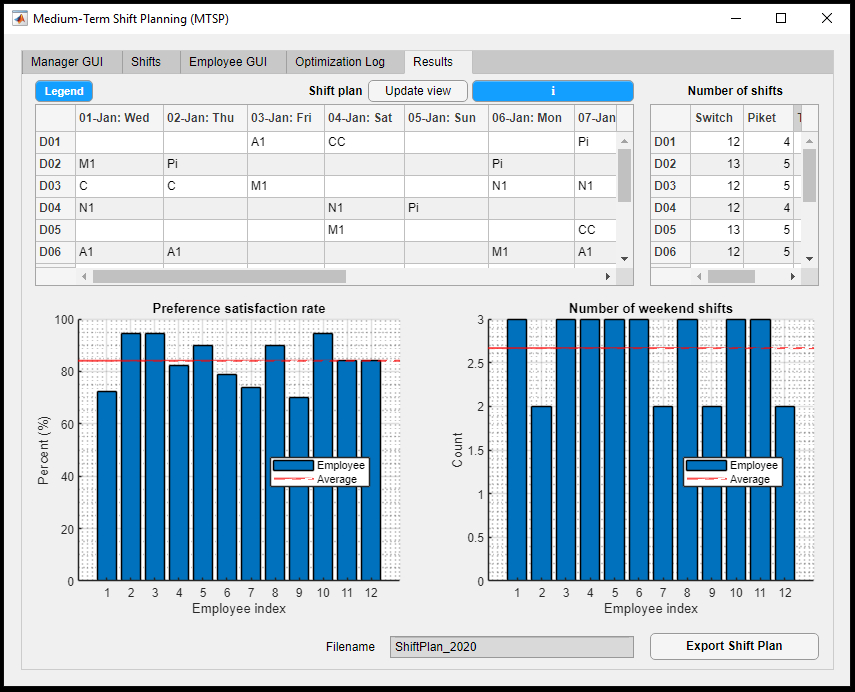}
    \caption{The software tool displays the optimized roster and basic statistics about it. In this example, the employees perform almost the same number of shifts overall as well as on weekends (to limit the size of the screenshot, the tabular results at the top are only visible for 6 of the 12 employees). Note that 85\% of the employee preferences are satisfied on average, with the lowest satisfaction rate being 70\% for employee 9. The solution can be exported to Excel for storage and further processing.}
    \label{fig:ResultsTab}
\end{figure}

\section{Use cases}\label{UseCases}
\subsection{Value for the company}
We packaged our model and algorithms for employee rostering as a software tool, whose design and functionalities are depicted in Figures \ref{fig:ManagerGUI} to \ref{fig:ResultsTab}. The software includes a GUI to accept manager and employee parameters, and display results. In this section, we present extensions to the basic implementation to meet the user needs at Swissgrid, which may also be relevant to other industries.

\subsection{Event-driven optimization}\label{EventDrivenOpt}
Employees can become unavailable unexpectedly for a number of reasons (e.g., illness). Absence leaves cover requirements under-satisfied, and others who are available must fill in. One strategy is to repeat the rostering from scratch with the new parameters. However, the new roster may be very different from the original roster and cause inconvenience to employees. Therefore, one should re-optimize the roster while minimizing its deviation from the original roster. 

We further found that small teams ($<$ 10 employees) at Swissgrid actually favored manual scheduling to an automatized solution because they have full control over their own work schedules. Instead, their main interest was not optimizing rosters from scratch, but re-optimizing existing rosters under new conditions communicated through change requests as many times as needed. This slightly modified implementation is called \textit{event-driven optimization}.



\begin{figure}
    \centering
    \includegraphics[width=1\textwidth]{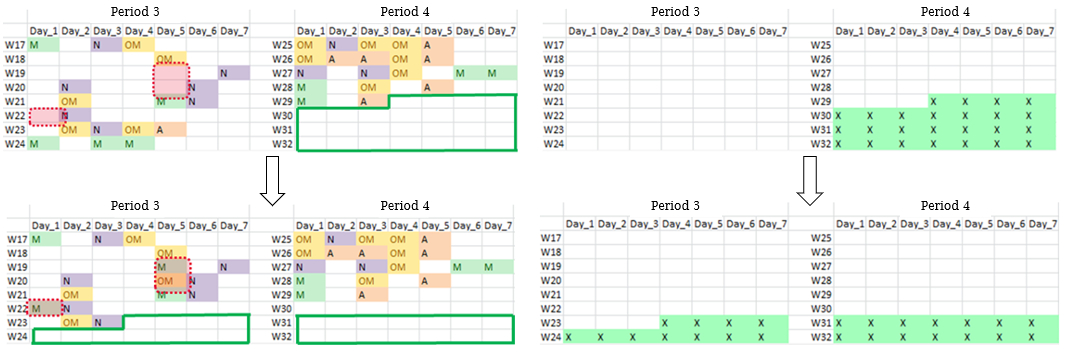}
    \caption{Example vacation change request (bottom right) and the re-optimized schedule for an employee from weeks 9 to 32 (bottom left). The original vacation parameter and schedule are shown on the top. The empty area enclosed in the green box indicates that the solution indeed does not violate vacation plans. The dashed red boxes highlight the changes in the re-optimized schedule compared with the original one, which are kept to a minimum.}
    \label{fig:EventDrivenOptEx}
\end{figure}

An example of vacation change request is presented in Figure \ref{fig:EventDrivenOptEx}. On the top, the original work schedule (left) and its corresponding vacation parameter (right) are given for an employee, while on the bottom are the the re-optimized schedule under the new vacation parameter. The shifts in weeks 23 and 24 in the original schedule that coincide with the new vacation plans are now removed in the re-optimized schedule. We further note that the degree of change is indeed minimal; Periods 1, 2, 5, 6, and 7, which are not shown, undergo no changes at all, and Period 3 is only slightly modified.

\subsection{Adaptive rolling horizon scheduling}
\textit{Adaptive rolling horizon scheduling} is a simple, but efficient technique to extend the planning horizon to one year, which would otherwise be impossible within reasonable time (e.g., 8 hours of work time) due to increased computational cost. Specifically, we divide one year into disjoint planning periods of several weeks each and optimize a roster in each period sequentially. As an example, we use seven disjoint 8-week periods.


From one period to the next, we implement an adaptive scheme which adjusts each employee's workload targets, for weekdays and weekends separately. The intuition behind the scheme is to increase (decrease) the workload targets for an employee whose cumulative workload has fallen short of (exceeded) his/her cumulative targets, by as much as the actual deviation of the cumulative workload from cumulative target. In so doing, we attempt to uniformize workload throughout the year on weekdays and weekends. 

\begin{figure}
    \centering
    \includegraphics[width=0.7\textwidth]{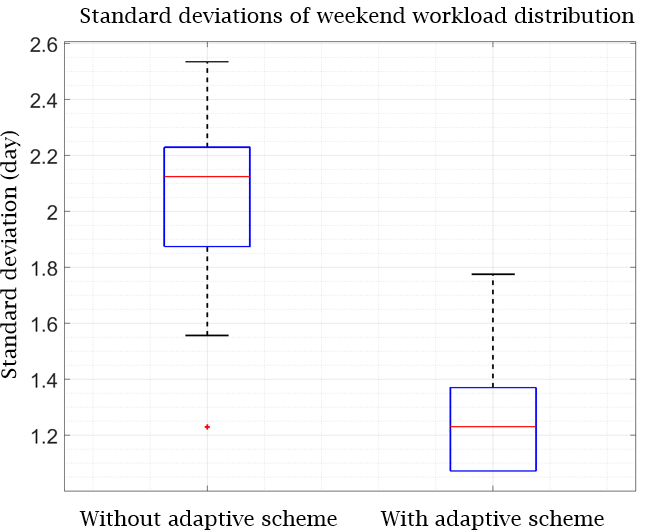}
    \caption{Comparison of the standard deviation of weekend workload distribution (in number of days) for two scenarios: with and without the adaptive scheme. The results are obtained by solving 10 randomized problems with 12 employees using the MILP-alone approach. For each of the randomized problems, the standard deviation is computed across the 12 dispatchers.}
    \label{fig:adaptscheme}
\end{figure}

In Figure \ref{fig:adaptscheme}, we present the standard deviations of employees' weekend workload from 10 randomized problems with 12 employees. A small standard deviation indicates that weekend shifts are distributed relatively equally. Figure \ref{fig:adaptscheme} shows that using the adaptive scheme produces smaller standard deviations with an average standard deviation of 1.2 days (compared to 2.1 days without the adaptive scheme). In other words, with the adaptive scheme most employees are within one Saturday or Sunday shift of each other over an entire year, but this increases to a full weekend without the adaptive scheme.

\subsection{Incorporation of company-defined work patterns}\label{CompanyPatterns}
\begin{figure}
    \centering
    \includegraphics[width=1\textwidth]{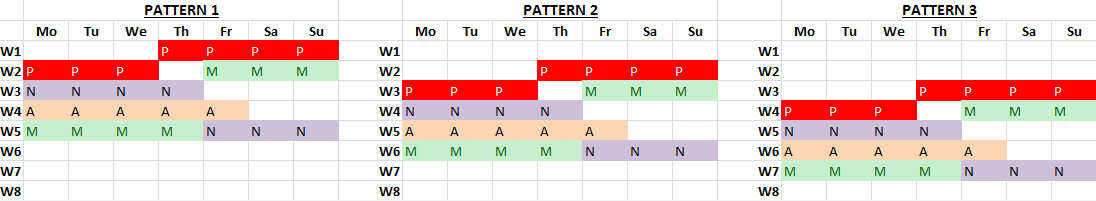}
    \caption{Example of a company-defined work pattern (left) and two of its weekly variations (middle and right) from the considered team at Swissgrid.}
    \label{fig:ExamplePatterns}
\end{figure}

Aside from employees' individual preferences, a company may wish to incorporate its organizational preference into the roster. The so-called \textit{company-defined work patterns} are imposed on all employees alike such that everyone has effectively the same schedule, different only in that it begins in different weeks for each employee. These work patterns may be defined for 8 weeks, which are then repeated for the rest of the year. Therefore, there may typically be only one unique work pattern, but a total of eight patterns are available since there are eight possible starting weeks. Figure \ref{fig:ExamplePatterns} depicts example work patterns in year 2020 from the considered team at Swissgrid.

\begin{figure}
    \centering
    \includegraphics[width=1\textwidth]{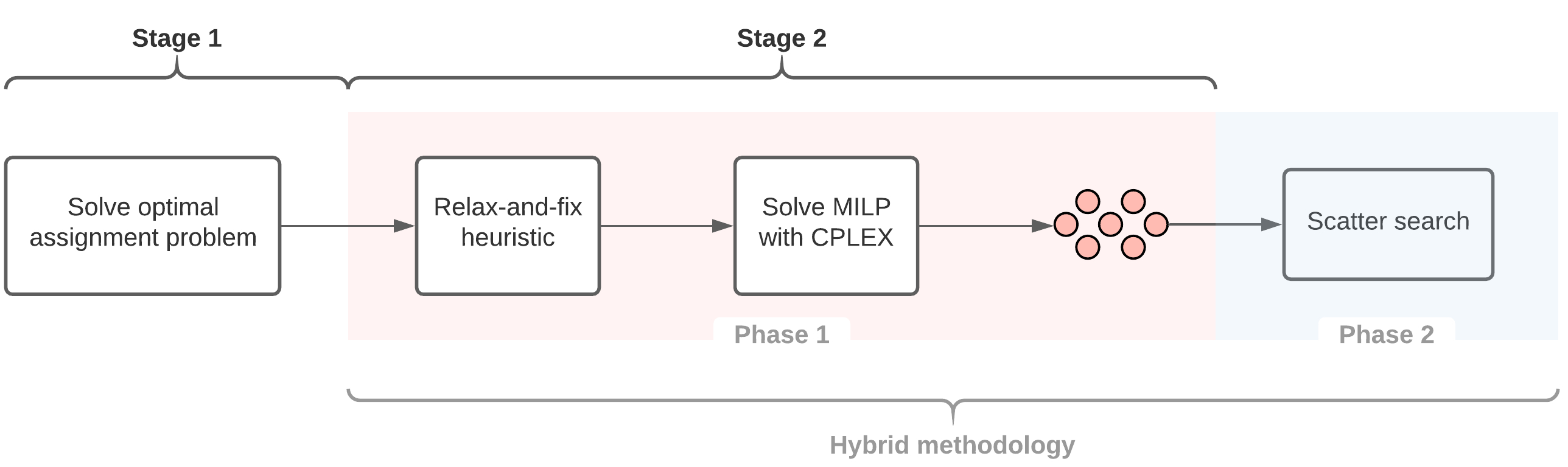}
    \caption{Flowchart of the two-stage MILP used to integrate company-defined work patterns in employee rostering. Phase 2 of the hybrid methodology is described in greater detail in Figure \ref{fig:Flowchart}.}
    \label{fig:two-stage MILP}
\end{figure}

Integrating company-defined work patterns is achieved by a two-stage MILP, as illustrated in Figure \ref{fig:two-stage MILP}.
The first-stage problem \textit{precedes} the relax-and-fix heuristic step in our hybrid methodology (see Figure \ref{fig:Flowchart}), and finds the optimal assignment of one work pattern per employee. It minimizes violation of employees' availability, vacation, and preferences by using non-negative slack and surplus variables in the constraints and minimizing their sum. If any one such variable is positive, this implies infeasibility, which is then corrected in the second-stage problem. The optimal solution to the first-stage problem, aptly called the \textit{company preference} parameter, is used in the second-stage problem, which is phase 1 of our hybrid methodology (see Figure \ref{fig:Flowchart}).

The second-stage problem is equal to the original employee rostering problem, but minimizes the violation of both individual and company preference parameters, as detailed in Appendix A. We strike a trade-off between the two preference-related objectives with a weight coefficient, $\gamma \in [0, 1]$. The closer $\gamma$ is to 0, the more priority is given to satisfying company preferences.

\begin{figure}
    \centering
    \includegraphics[width=1\textwidth]{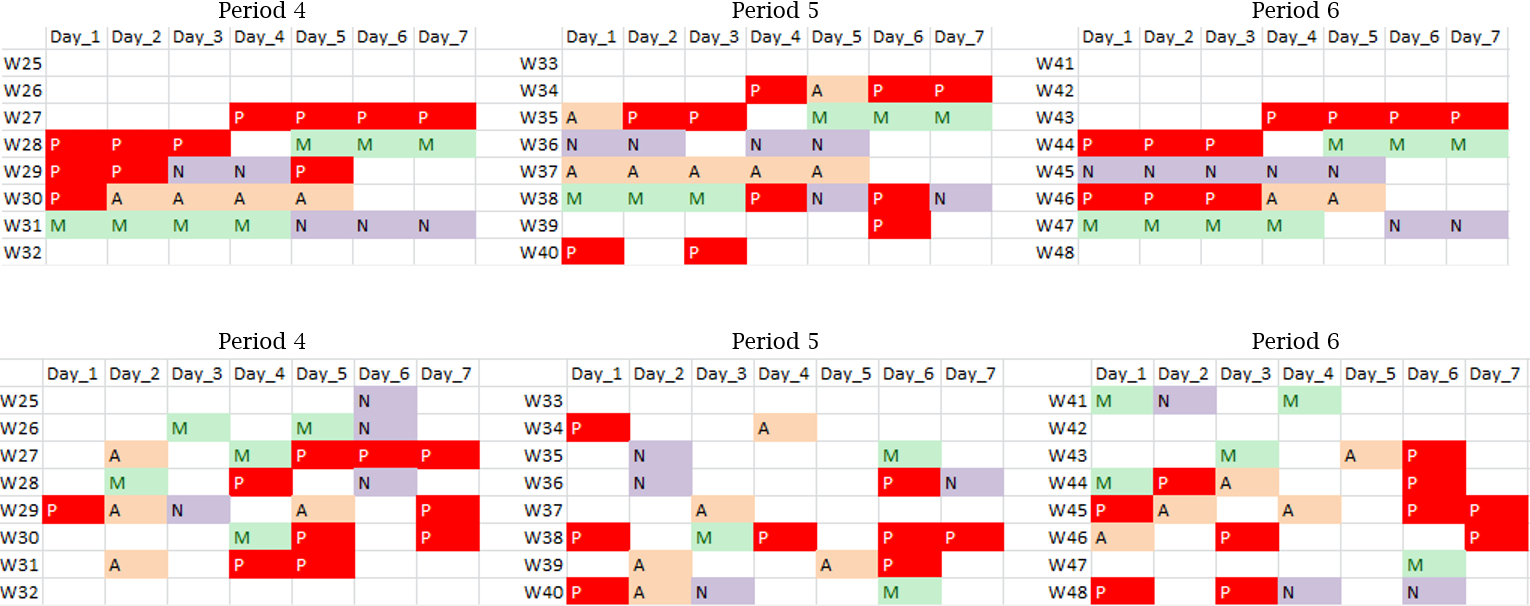}
    \caption{Top: Example schedule with a full regard for company-defined work patterns ($\gamma=0$). The schedule closely resembles the examples in Figure \ref{fig:ExamplePatterns}; however, perfect replication is impossible due to employees' own unavailability. Bottom: Example schedule with no regard for the work patterns ($\gamma=1$) for the same employee with identical parameters. The schedule is more scattered and does not follow the examples in Figure \ref{fig:ExamplePatterns} closely.}
    \label{fig:IncorporatePatterns}
\end{figure}

Figure \ref{fig:IncorporatePatterns} shows two use cases: rostering with $\gamma=100\%$ (top) and $\gamma=0\%$ (bottom). Note that there is a significant difference in the schedules' appearance despite being the same employee with identical parameters. Indeed, full priority over the work patterns produces a schedule with a close affinity to the examples in Figure \ref{fig:ExamplePatterns}. By tuning $\gamma$, we can create a schedule which is balanced between these two extreme attitudes to scheduling.

\subsection{Additional use cases}
Two additional extensions include \textit{vacation co-optimization} and \textit{staffing level optimization}, which are only briefly discussed here. The former is a functionality that corrects infeasibility from significant overlap among employees' vacation; it makes minimal adjustments while preventing the vacation time from breaking into undesired, disjoint time periods. Staffing level optimization is a functionality that uses the employee rostering model as a simulator to discover the optimal team size and license mix for the various shift types under competing objectives.

\section{Conclusion and future work}\label{Conclusion}
In this work, we presented a new hybrid methodology to optimize employee rostering and applied it for a team at Swissgrid. The hybrid methodology combines an MILP model with a metaheuristic called scatter search, and could solve many problems involving 24 to 72 employees by an order of magnitude faster than the MILP when solved by CPLEX alone. Additional practical advantages were observed such as considerably slower growth in computational cost with respect to the number of employees and producing more reliable and robust solutions under different randomized parameters. 

Being an industry project, our model must reflect the needs of the stakeholders. Among these needs, we identified that our tool must optimize adjustments to pre-computed rosters when problem parameters change unexpectedly, incorporate company-defined work patterns with a user-defined priority level, and scale up the planning horizon to one year in an efficient manner. Simulations illustrated that our tool satisfies all of these needs.

We believe that scheduling software tools with scalable solution engines and a comprehensive application layer, such as the methodology and tool presented in this paper, are required to achieve a widespread adoption of automated rostering solutions in the industry. Among other things, future work on the methodology side will target stochastic extensions of the scheduling optimization problem, as well as computational improvements by means of decomposition techniques. In terms of new software functionalities, we will focus on an enhanced user interface and modeling language that allows the user to input custom scheduling constraints in a straightforward way.
%

\section*{Acknowledgement(s)} 
The authors would like to thank Dr. Marek Zima, Stéphane Duarte Dätwyler, Joël Egg and Bastien Grand from Swissgrid, and Prof. Dr. Daniel Kuhn and Dirk Lauinger from EPFL for fruitful discussions and feedback.






\bigskip

\appendix

\section{Mathematical formulation}
\subsection{Notation}
\subsubsection{Parameters and sets}
Table \ref{tab:Parameters} summarizes all parameters involved in the mathematical formulation. The index $e$ is used to refer to each employee; the index $j$ refers to each \textit{time block}, which is a pre-defined 8-hour period within a day such as 6:00 - 14:00; lastly, the index $k$ is used to refer to each shift type.
\begin{table}
    \tbl{Problem parameters and their descriptions.}{
    \begin{tabular}{c p{12cm}} \toprule
        Parameter & Description \\
    \midrule
    $w$ & Number of weeks in the planning horizon \\
    $n$ & Number of employees \\
    $s$ & Number of shift types \\
    $m$ & Number of time blocks in the planning horizon, which is equal to $21w$. (Each week has seven days, and each day has three 8-hour time blocks; hence 21.) \\
    $\overline{h}$ & Maximum number of shifts per week \\
    $\underline{h}$ & Minimum number of shifts per week \\
    $r_{min}$ & Minimum number of rest days per year \\
    $l_{min}$ & Minimum number of rest Sundays per year \\
    $A$ & \{0,1\}$^{n\times m}$ matrix parameter representing employee availability; $A_{ej} = 1$ means employee $e$ is available during time block $j$. \\
    $V$ & \{0,1\}$^{n\times m}$ matrix parameter representing employee vacation; $V_{ej} = 1$ means employee $e$ is on vacation during time block $j$. \\
    $P$ & $n\times m$ matrix parameter representing employee preferences; $P_{ej}$ = 1 means employee $e$ prefers to work during time block $j$ whereas 0 means he does not prefer a shift. A missing (null) entry indicates no preference. \\
    $D$ & $\mathbb{Z}_+^{m\times s}$ matrix parameter representing the cover requirements; $D_{jk} = 1$ means one employee is required during time block $j$ for shift type $k$. \\
    $T_{ek}$ & Scalar representing the end-of-horizon workload target for employee $e$ for shift type $k$. \\
    $G_{ek}$ & Scalar representing the end-of-horizon weekend workload target for employee $e$. \\
    \bottomrule
    \end{tabular}}
    \label{tab:Parameters}
\end{table}

Table \ref{tab:Sets} outlines a list of sets involved in the mathematical formulation. For an arbitrary matrix $A$, the notation $A_{e*}$ represents the $e^{th}$ row of the matrix $A$. Similarly, the $j^{th}$ column of the matrix is denoted by $A_{*j}$.
\begin{table}
    \tbl{Index sets and their descriptions.}{
    \begin{tabular}{c p{12cm}} \toprule
        Set & Description \\
    \midrule
    $\mathcal{P}_e$ & Set of non-null indices in $P_{e*}$ for every $e \in \{1, ..., n\}$. \\
    $\mathcal{N}_k$ & Set of employees without the necessary license to perform shift type $k$, defined for every $k \in \{1, ..., s\}$. \\
    $\mathcal{S}$ & Set of indices which correspond to Sunday time blocks; $\mathcal{S} \subset \{1, ..., m\}$. \\
    $\mathcal{W}$ & Set of indices which correspond to weekend time blocks; $\mathcal{W} \subset \{1, ..., m\}$. \\
    \bottomrule
    \end{tabular}}
    \label{tab:Sets}
\end{table}

\subsubsection{Decision variables}
A three-dimensional binary (0/1) decision variable $X$ represents the employee assignments: $X_{ejk} = 1$ if employee $e$ is assigned during time block $j$ to shift type $k$; 0 otherwise. Continuous auxiliary variables are also introduced to linearize $L_1$ and $L_\infty$ norms in the objective function and constraints.

\subsection{Objectives and constraints of basic implementation}
The overall objective function is defined as follows:
\begin{equation}
    \min_X \quad F(X) = \sum^3_{i=1}\lambda_i\left[\theta_if_i(X) + (1-\theta_i)\overline{f}_i(X)\right] \label{Obj}
\end{equation}

Where the individual objective terms are:
\begin{itemize}
    
    
    \item Deviation of employees' workload from their targets
    \begin{subequations}
    \begin{align}
        f_1(X) &= \sum^s_{k=1}\sum^n_{e=1}\left|T_{ek} - \sum^m_{j=1}X_{ejk}\right|\label{first_L1}\\
        \overline{f}_1(X) &= \sum^s_{k=1}\max_e\left|T_{ek} - \sum^m_{j=1}X_{ejk}\right|\label{first_Linf}
    \end{align}
    \end{subequations}
    
    \item Deviation of employees' weekend workload from their weekend workload targets
    \begin{subequations}
    \begin{align}
        f_2(X) &= \sum^s_{k=1}\sum^n_{e=1}\left|G_{ek} - \sum_{j\in\mathcal{W}} X_{ejk}\right| \label{second_L1} \\
        \overline{f}_2(X) &= \sum^s_{k=1}\max_e\left|G_{ek} - \sum_{j\in\mathcal{W}} X_{ejk}\right| \label{second_Linf}
    \end{align}
    \end{subequations}
    
    \item Violation of employees' work preferences
    \begin{subequations}
    \begin{align}
        f_3(X) &= \sum^n_{e=1}\sum_{j\in\mathcal{P}_e}\left|\sum^s_{k=1}X_{ejk} - P_{ej}\right| \label{prefobj} \\
        \overline{f}_3(X) &= 0 
    \end{align}
    \end{subequations}
    
\end{itemize}

where the weighting factors $\theta_i$ and $\lambda_i$ lie in [0, 1] for every $i \in \{1,2,3\}$. $f_1(X), f_2(X)$ and $f_3(X)$ are $L_1$ norms, whereas $\overline{f}_1(X)$ and $\overline{f}_2(X)$ are $L_\infty$ norms. \newline

All other requirements of employee rostering are implemented as hard constraints:
\begin{itemize}
\setlength\itemsep{-0.5em}
    \item Employees must not be assigned to more than one shift type at any time.
    \begin{align}
        \sum^s_{k=1}X_{ejk} &\leq 1, \qquad \forall e=1,...,n, j=1,...,m \label{con1}
    \end{align}
    
    \item Employees are assigned to a shift type only if they have the license for it.
    \begin{align}
        X_{ejk} &= 0, \qquad \forall e\in\mathcal{N}_k, j=1,...,m, k=1,...,s
    \end{align}
    
    \item Both ``P" (on-call duty) and ``OM" (outage planning) last 24 hours. 
    
    For a particular index $k$ which corresponds to ``P" or ``OM":
    \begin{align}
        X_{e,3(t-1)+1,k} = X_{e,3(t-1)+2,k} = X_{e,3(t-1)+3,k}, \qquad \forall e\notin \mathcal{N}_k, t=1,...,\frac{m}{3}
    \end{align}
    
    \item Employees must not be assigned to any shifts when they are unavailable.
    \begin{align}
        X_{e*k} \leq A_{e*}, \qquad \forall e=1,...,n, k=1,...,s
    \end{align}
    
    \item Employees on vacation must not be assigned to any shifts during those times.
    \begin{align}
        X_{e*k} \leq (1-V_{e*}), \qquad \forall e=1,...,n, k=1,...,s \label{vaccon}
    \end{align}
    
    \item Employees must have at least 16 hours of rest between any two switching operations shifts.
    
    For a particular index $k$ which corresponds to switching operations:
    \begin{align}
        \sum^{t+2}_{j=t}X_{ejk} \leq 1, \qquad \forall e\notin \mathcal{N}_k, t=1,...,m-2
    \end{align}
    
    \item Maximum allowed number of switching operations shifts per week per employee.
    
    Again, for a particular index $k$ which corresponds to switching operations:
    \begin{align}
        \sum^{21(t-1)+21}_{j=21(t-1)+1}X_{ejk} \leq \overline{h}, \qquad \forall e\notin \mathcal{N}_k, t=1,...,w
    \end{align}
    where 21 comes from the fact that there are three (8-hour) time blocks per day and seven days per week. Minimum allowed number of shifts per week is implemented similarly:
    \begin{align}
        \sum^{21(t-1)+21}_{j=21(t-1)+1}X_{ejk} \geq \underline{h}, \qquad \forall e\notin \mathcal{N}_k, t=1,...,w
    \end{align}
    
    \item Minimum required number of rest Sundays per year.
    \begin{align}
        w-\sum^s_{k=1}\sum_{j\in\mathcal{S}}X_{ejk} \geq l_{min}, \qquad \forall e=1,...,n
    \end{align}
    
    \item Cover requirements must be satisfied exactly.
    \begin{align}
        \sum^n_{e=1}X_{ejk} = D_{jk}, \qquad \forall j=1,...,m, k=1,...,s \label{con10}
    \end{align}
    
    \item Minimum required number of rest days per year.
    
    Counting rest days amounts to counting instances of consecutive 24-hour periods throughout the planning horizon. Within a mathematical program, this is achieved using continuous and integer auxiliary variables in the constraints.
\end{itemize}

To solve the MILP within the hybrid methodology we used MATLAB and ILOG CPLEX 12.6.1.0 in the distribution available in TOMLAB 8.1.

\subsection{Incorporation of company-defined work patterns}
In the two-stage MILP used to incorporate company-defined work patterns (see Section \ref{CompanyPatterns}), the second-stage problem admits the optimal solution of the first-stage problem and uses it as a parameter in a slightly altered objective function:
\begin{equation}
    \min_X \quad F^c(X) + \lambda_3\theta_3 \left[ \gamma f_3(X) + (1-\gamma)f_4(X)  \right] \label{CompanyPattern_Obj}
\end{equation}
where the individual terms are described as follows:
\begin{itemize}
    \item $F^c(X)$ is equal to the objective function (\ref{Obj}) which sums up to $i=2$, not 3.
    \begin{equation}
        F^c(X) = \sum^2_{i=1}\lambda_i\left[\theta_if_i(X) + (1-\theta_i)\overline{f}_i(X)\right]
    \end{equation}
    \item $f_3(X)$ is equal to objective (\ref{prefobj}) and quantifies employee preference violation.
    \item The new objective term $f_4(X)$ quantifies the deviation of employees' schedules, $X$, to their assigned company-defined work patterns, denoted $X^c$.
    \begin{equation}\label{f4_obj}
        f_4(X) = \sum^s_{k=1} \sum^n_{e=1} \sum^m_{j=1} \left| X^c_{ejk} - X_{ejk} \right|
    \end{equation}
    \item The closer the weight coefficient $\gamma \in [0,1]$ is to 0, the more priority is given to satisfying company-defined work patterns.
\end{itemize}

\end{document}